\documentclass[lettersize,journal]{IEEEtran}
\usepackage{amsmath,amsfonts}
\usepackage{algorithmic}
\usepackage{algorithm}
\usepackage{multirow}
\usepackage{multicol}
\usepackage{array}
\usepackage[caption=false,font=normalsize,labelfont=sf,textfont=sf]{subfig}
\usepackage{textcomp}
\usepackage{hyphenat}
\usepackage{stfloats}
\usepackage{url}
\usepackage{verbatim}
\usepackage{graphicx}
\usepackage{tikz} 
\usepackage{cite}
\usepackage{xcolor}
\usepackage{soul}
\usepackage{array}
\usepackage{afterpage} 
\usepackage{booktabs} 

\begin{document}

\title{AV-Lip-Sync+: Leveraging AV-HuBERT to Exploit \\ Multimodal Inconsistency for Deepfake Detection of Frontal Face Videos}

\author{Sahibzada Adil Shahzad, Ammarah Hashmi, Yan-Tsung Peng, Yu Tsao,~\IEEEmembership{Senior Member,~IEEE}, Hsin-Min Wang,~\IEEEmembership{Senior Member,~IEEE}
\thanks{Sahibzada Adil Shahzad is with the Social Networks and Human-Centered Computing Program, Taiwan International Graduate Program, Academia Sinica, Taipei 11529, Taiwan, and also with the Department of Computer Science, National
Chengchi University, Taipei 11605, Taiwan. (e-mail: adilshah275@iis.sinica.edu.tw).

Ammarah Hashmi is with the Social Networks and Human-Centered Computing Program, Taiwan International Graduate Program, Academia Sinica, Taipei 11529, Taiwan, and also with the Institute of Information Systems and Applications, National Tsing Hua University, Hsinchu 30013, Taiwan. (e-mail: hashmiammarah0@gmail.com).

Yan-Tsung Peng is with the Department of Computer Science, National Chengchi University, Taipei, Taiwan. (e-mail: ytpeng@cs.nccu.edu.tw)

Yu Tsao is with the Research Center for Information Technology Innovation, Academia Sinica, Taipei 11529, Taiwan. (e-mail:
yu.tsao@citi.sinica.edu.tw).

Hsin-Min Wang is with the Institute of Information Science, Academia Sinica, Taipei 11529, Taiwan. (e-mail: whm@iis.sinica.edu.tw)

}}



\maketitle
\begin{abstract}
Multimodal manipulations (also known as audio-visual deepfakes) make it difficult for unimodal deepfake detectors to detect forgeries in multimedia content. To avoid the spread of false propaganda and fake news, timely detection is crucial. The damage to either modality (i.e., visual or audio) can only be discovered through multimodal models that can exploit both pieces of information simultaneously. However, previous methods mainly adopt unimodal video forensics and use supervised pre-training for forgery detection. This study proposes a new method based on a multimodal self-supervised-learning (SSL) feature extractor to exploit inconsistency between audio and visual modalities for multimodal video forgery detection. We use the transformer-based SSL pre-trained Audio-Visual HuBERT (AV-HuBERT) model as a visual and acoustic feature extractor and a multi-scale temporal convolutional neural network to capture the temporal correlation between the audio and visual modalities. Since AV-HuBERT only extracts visual features from the lip region, we also adopt another transformer-based video model to exploit facial features and capture spatial and temporal artifacts caused during the deepfake generation process. Experimental results show that our model outperforms all existing models and achieves new state-of-the-art performance on the FakeAVCeleb and DeepfakeTIMIT datasets. 
\end{abstract}

\begin{figure}
\centering
\includegraphics[width=\linewidth]{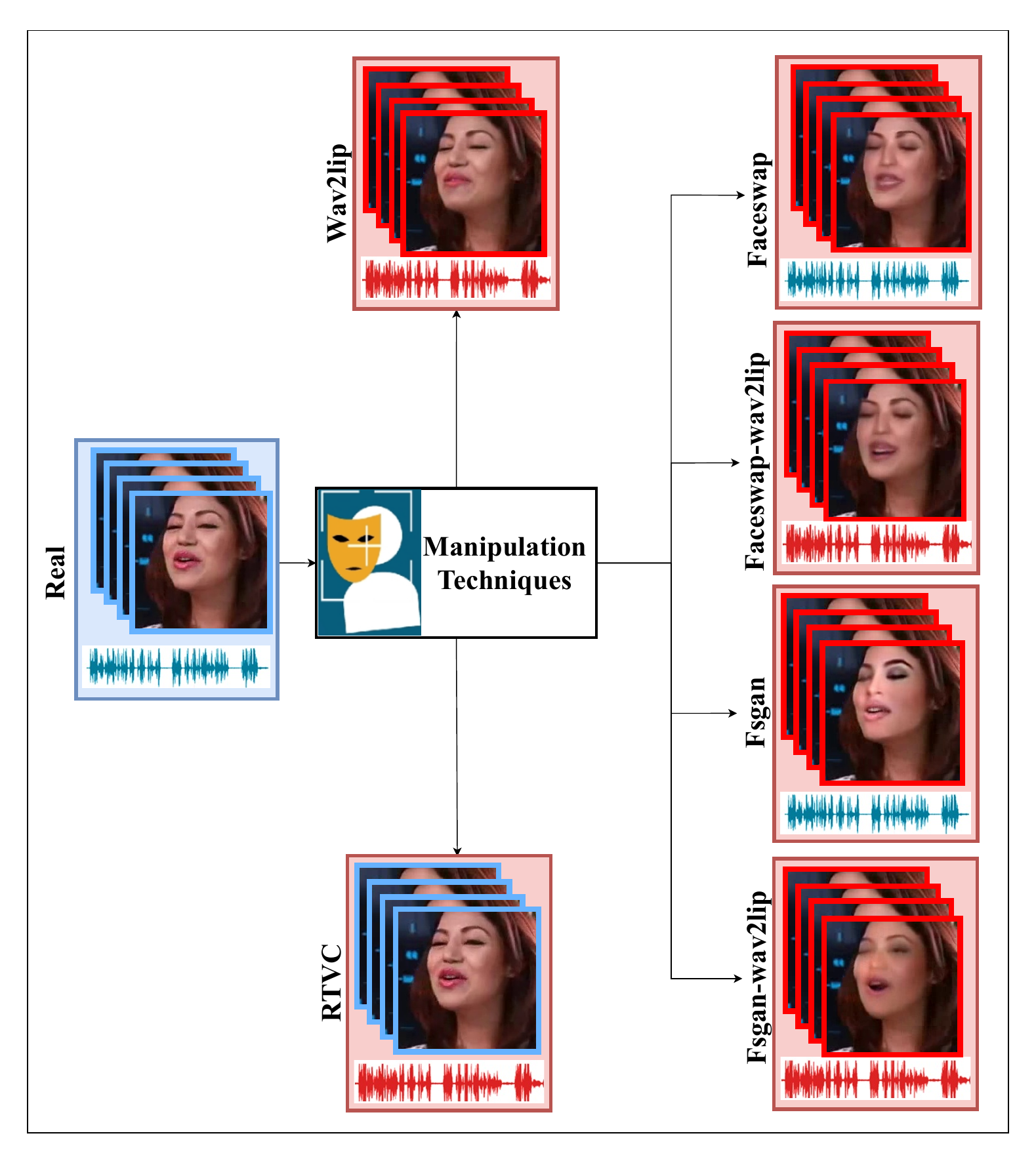} 
\caption{Illustration of various deepfake manipulation techniques applied to a real audio-visual sample. The real sample (left) highlighted by the dark blue border contains the original video frames and corresponding audio waveform. The manipulated (fake) samples highlighted by dark red borders are generated using Wav2Lip, Faceswap, Faceswap-wav2Lip, Fsgan, Fsgan-wav2Lip, and RTVC (Real-Time Voice Cloning). The video frames highlighted by blue borders represent real frames, while the video frames highlighted by red borders represent manipulated (fake) frames. The blue waveforms represent real audio, while the red waveforms represent manipulated (fake) audio.}
\label{fig_1}
\vspace{-4mm}
\end{figure}

\section{Introduction}
With smartphones, social networks, and the high-speed internet, it is now available at one’s fingertips to capture, upload, and share content without any delays or fees. However, this convenience makes the spread of deepfakes a heavy social cost raising major social and ethical issues~\cite{vaccari2020deepfakes, choras2021advanced}. The term ``deepfake'' encompasses synthetic media such as images, videos, audio, and text. While there are many benefits to generating content through artificial intelligence (AI) technology, it also has a dark side.  AI-generated content is often used for unethical objectives and malicious purposes, including disinformation, pornography, fraudulent activities, and political defamation~\cite{ray2021disinformation, choras2021advanced, figueira2017current, vaccari2020deepfakes}. Social networks~\cite{narayan2022desi} are conduits for this type of manipulated content and often do not have filtering mechanisms to prevent its rapid spread.

Deepfakes exist in various modalities, each posing distinct challenges. Text deepfakes refer to seemingly real text information manipulated or generated by AI. Examples include fabricated news articles~\cite{zellers2019defending}, deceptive online reviews~\cite{yao2017automated}, and synthetic text that incites violence~\cite{mcguffie2020radicalization}. Audio deepfakes involve the acoustic manipulation of audio using voice conversion and text-to-speech techniques (such as WaveNet~\cite{vandenoord16_ssw}, WaveGlow~\cite{prenger2019waveglow}, MelNet~\cite{vasquez2019melnet}, and Tacotron~\cite{wang2017tacotron}). SV2TTS~\cite{jia2018transfer} is a real-time voice cloning tool that synthesizes fabricated audio content without altering visual content, such as video frames. Among the various forms of deepfake media, video deepfakes are the most prevalent. These AI-generated videos realistically superimpose one person's face onto another, modifying expressions or mimicking speech patterns. Faceswap~\cite{korshunova2017fast} and FSGAN~\cite{nirkin2019fsgan} are common techniques for visual deepfake manipulation.

\par Advances in deepfake generation technology have made the task of facial forgery detection significantly challenging. Audio-visual deepfake techniques have reached new levels of realism with this advancement. The realistic nature of audio-visual deepfake videos stems from the simultaneous manipulation of both audio and visual modalities and the consistency between synthesized speech and facial movements, making traditional forgery detection methods less effective. Humans often fail to distinguish real from manipulated videos, yet remain overconfident in their judgments~\cite{kobis2021fooled, diel2024human}. Similarly, recent multimodal large language models (LLMs) such as ChatGPT~\cite{jia2024can} demonstrate limited effectiveness in multimedia forensics and struggle with highly realistic manipulations. These challenges require more advanced detection strategies that go beyond simple frame-based analysis and the development of specialized tools and algorithms designed specifically for multimedia forensics tasks. To solve this problem, the forensics research communities in the fields of image, video, and audio have specially designed algorithms to detect manipulation in videos. The image and video forensics research community focuses primarily on detecting visual forgeries~\cite{haliassos2021lips, nguyen2019capsule, lutz2021Deepfake, chollet2017xception,  afchar2018mesonet}, while the audio forensics research community has developed detection systems to detect acoustic manipulation~\cite{wang2015relative, todisco2016new, patel2015combining}. Due to their unimodal nature, these detectors fail when the manipulated modality is not seen during training. Spoofed audio can evade visual deepfake detectors, while audio deepfake detectors cannot catch visual deepfakes. 

\par Recently, inspired by human's ability to process audio and visual signals simultaneously to perceive the world~\cite{scheliga2023neural}, the field of multimedia forensics has turned to the development of multimodal systems for effective deepfake detection. Humans instinctively combine auditory and visual signals to assess the authenticity of content, often relying on the synchronization of speech and facial movements. Likewise, utilizing multimodal data such as audio and visual modalities has emerged as a promising approach to improve the accuracy and robustness of video forgery detection~\cite{zhang2019multimodal}. Therefore, we choose to integrate audio and visual information into our bimodal approach to identify audio-visual deepfakes. To further facilitate this audio-visual deepfake detection technology, various pre-trained models were fine-tuned on deepfake video datasets to detect manipulation.

\par Pre-trained self-supervised-learning (SSL) models have recently emerged and achieved success in various downstream tasks. Audio-Visual HuBERT (AV-HuBERT)~\cite{shi2021learning} is an SSL-based audio-visual representation learning model that achieves state-of-the-art performance in lip reading, audio-visual speech enhancement~\cite{chern2023audio}, audio-visual speech separation~\cite{chern2023audio}, and audio-visual speech recognition~\cite{shi22_interspeech}. Motivated by its state-of-the-art performance in multiple downstream tasks, we leverage AV-HuBERT for feature extraction to capture the inconsistency between the mouth region of interest and the corresponding audio modality for the downstream task of audio-visual deepfake detection. 
Considering that lip feature-based deepfake detectors may fail when the lip region is not manipulated or only slightly manipulated, to enhance our audio-visual forgery detection model, we also employ Video Vision Transformer (ViViT)~\cite{arnab2021vivit} as the face encoder to exploit whole-face features to assist our proposed audio-visual deepfake detector.
By integrating powerful audio-visual representations, speech-lip synchronization features, and spatiotemporal facial features, the proposed system achieves state-of-the-art performance on the FakeAVCeleb~\cite{khalid2021fakeavceleb} and DeepfakeTIMIT~\cite{korshunov2018deepfakes} datasets. 

This work focuses on deepfake detection of frontal face videos with speech in the audio track. Such deepfake videos of celebrities are widely circulated on social media platforms to spread misinformation (or disinformation) or tarnish a person's reputation, causing serious harm to society. Our main contributions are as follows.

\begin{itemize}
    \item We propose AV-Lip-Sync+, a novel audio-visual deepfake detection model that combines self-supervised speech-lip synchronization features, audio-visual embeddings, and facial representations to capture multimodal inconsistencies and spatiotemporal artifacts.
    
    \item We leverage a transformer-based architecture to improve temporal modeling and cross-modal alignment, surpassing conventional CNN-based approaches in capturing subtle deepfake cues.

    \item We introduce a dedicated face encoder to enhance the detection performance on full-face manipulations (e.g., Faceswap and FSGAN), and evaluate the model's robustness under generalization and partial face occlusion scenarios.

    \item Extensive experiments on multiple benchmark datasets, including FakeAVCeleb, DeepfakeTIMIT, and DFDC, demonstrate that our approach consistently outperforms existing state-of-the-art methods in terms of accuracy and robustness.   
\end{itemize}

\section{Related Work} In this section, we first briefly review common deepfake video generation techniques, and then introduce state-of-the-art methods for deepfake detection.

\subsection{Deepfake Generation} 

Deepfake manipulation comes in many forms, with visual forgery~\cite{korshunova2017fast, chen2020simswap, gao2021information,thies2019deferred, li2019faceshifter, thies2016face2face} being the most common, which involves using deepfake generation models such as Faceswap~\cite{korshunova2017fast}, FSGAN~\cite{nirkin2019fsgan}, and wav2lip~\cite{prajwal2020lip} to manipulate the entire face, facial expressions, or lip movements. Traditional deepfake techniques are limited to visual forgery, leaving the audio unaltered. The evolution of synthetic media technology has significantly improved the realism of deceptive content, among which audio-visual deepfakes enable the integration of visual and auditory alterations. Techniques like Faceswap-wav2lip and Fsgan-wav2lip simultaneously manipulate face and lip movements and align the latter with the audio track. Fig.~\ref{fig_1} shows real and various audio-visual fake video samples from the FakeAVCeleb dataset~\cite{khalid2021fakeavceleb}.
The focus of our study is on detecting audio-visual deepfakes.

\subsection{Deepfake Detection}  High-quality AI-generated content is useful in many ways, but it also comes at a cost, and timely detection is crucial to avoid any harm to society~\cite{ray2021disinformation}. To avoid the spread of misinformation and disinformation and to protect the reputations and privacy of individuals, we need automated methods to promptly detect deepfake content in our widespread digital world. Academia and industry have made considerable progress in using deep learning-based methods to detect forged multimedia content. These deep learning-based deepfake detection methods can be roughly divided into two major categories: unimodal and multimodal methods. 

Unimodal forgery detection models are specifically designed to detect forgery in one modality (video or audio) and rely only on the corresponding modality to identify forgery. Video forgery detectors can be divided into three types \cite{lyu2020deepfake}: physiological, visual artifact-based, and high-level feature-based methods. In physiological methods, researchers have exploited abnormal eye blinks~\cite{li2018ictu} and incoherent head poses~\cite{yang2019exposing}. Visual artifact-based methods analyze anomalies and irregularities such as unnatural facial movements, illumination variation, blended face boundaries, and misalignment of video content. FInfer~\cite{hu2022finfer} is a frame inference-based detection framework to solve the problem of high-visual-quality deepfake detection. ICM~\cite{hu2022dynamic} is a deepfake detection model that captures dynamic inconsistencies between visual frames in deepfake videos. GFA-CNN~\cite{tu2020learning} is proposed to learn identity-aware and generalizable features for face anti-spoofing tasks. 
High-level-feature-based deepfake detectors extract high-level features that are immune to video processing (e.g., compression). Lipforensics~\cite{haliassos2021lips} is an example of a high-level feature-based deepfake detector that leverages a lip-reading-based model to detect abnormal lip movements for video deepfake detection.

Video forgery detection models have achieved excellent results, thanks to various rich datasets available for model training, such as DeepfakeTIMIT~\cite{korshunov2018deepfakes}, UADVF~\cite{yang2019exposing}, FaceForensics++~\cite{rossler2019faceforensics++}, Celeb-DF~\cite{li2020celeb}, DFDC~\cite{dolhansky2020deepfake}, DeeperForensics~\cite{jiang2020deeperforensics}, and the recently released multimodal FakeAVCeleb dataset~\cite{khalid2021fakeavceleb}. Well-known unimodal fake video detection models include Capsule Forensics~\cite{nguyen2019capsule}, HeadPose~\cite{lutz2021Deepfake}, Xception~\cite{chollet2017xception}, LipForensics~\cite{haliassos2021lips}, Meso-4 and MesoInception-4~\cite{afchar2018mesonet}.

On the other hand, to trick automatic speaker verification systems, attackers can develop audio spoofing attacks or replay attacks using only a few minutes of a person's recorded speech, which makes these systems vulnerable. Similar to video deepfakes, audio spoofing attacks are a major challenge that must be solved. In response to audio spoofing attacks on automatic speaker verification systems, the audio forensics community has proposed various traditional and deep learning-based methods~\cite{khan2022voice}. Most traditional systems use short-term power spectrum, short-term phase spectrum, and long-term spectral features as front-end features. Backend classifiers are based on traditional machine learning models or ensemble models. Deep learning-based models can be roughly divided into multi-pass, end-to-end, and ensemble models. Using different hand-crafted acoustic features or raw waveforms, audio spoofing detection methods ~\cite{wang2015relative, todisco2016new, patel2015combining} can differentiate between genuine speech and spoofed speech. A step forward from whole-utterance spoof detection, H-MIL~\cite{zhu2023local} employs a multiple instance learning approach for partially spoofed speech detection. 

Recently, deepfake technology has expanded from unimodal manipulation to multimodal manipulation (such as audio-visual manipulation). Audio-visual deepfakes involve simultaneously manipulating the facial features and speech patterns of a subject in a video, making them more intricate and less detectable. Unimodal deepfake video and spoofed audio detectors are insufficient to detect these audio-visual manipulations. Despite their effectiveness in certain cases, these unimodal techniques often lack the ability to identify crossmodal inconsistencies. To solve this problem, the multimedia forensics community has proposed several deep learning-based methods to capture unimodal or multimodal manipulations. To effectively identify audio-visual deepfakes and reduce their misuse, advanced detection techniques that combine multimodal fusion, temporal consistency analysis, and deep learning-based anomaly detection are needed, all of which face considerable challenges.

Few studies have addressed this problem through multimodal methods that exploit faces as visual features and mel-frequency cepstral coefficients (MFCCs) as acoustic features~\cite{chugh2020not}, or detect forged videos based on facial and speech emotions~\cite{mittal2020emotions}. To capture the intrinsic synchronization between visual and acoustic modalities, Facebook AI \cite{zhou2021joint} built a sync-stream by connecting video and audio network feature representations through intra-attention (self-attention) and inter-attention mechanisms within and across the video and audio modalities. 
In~\cite{shahzad2022lip}, speech-lip synchronization features based on the difference between the extracted lip sequence and the synthetic lip sequence are used to detect audio-visual deepfakes. This model requires the wav2lip~\cite{hegde2021visual} module to generate lip sequences from audio, which increases model complexity, training, and inference time.
AVFakeNet audio-visual forgery detection method~\cite{ilyas2023avfakenet} uses Swin Transformer as the feature extraction module. 
In~\cite{hashmi2022multimodal}, a multimodal deepfake detector based on ensemble learning is introduced to leverage multiple learners and make decisions based on hard voting. 
In its subsequent extended model in~\cite{hashmi2023avtenet}, three separate models, namely audio network, video network, and audio-visual network, are all built on pre-trained transformer-based foundation models.
The main disadvantage of ensemble learning is its time-consuming process due to the involvement of multiple training models. 

The field of multimodal forgery detection is still less explored, and further research is needed to determine how to effectively use audio-visual information to detect forgery in any modality of multimedia content.

\begin{figure*}[t!]
  \includegraphics[width=\textwidth]{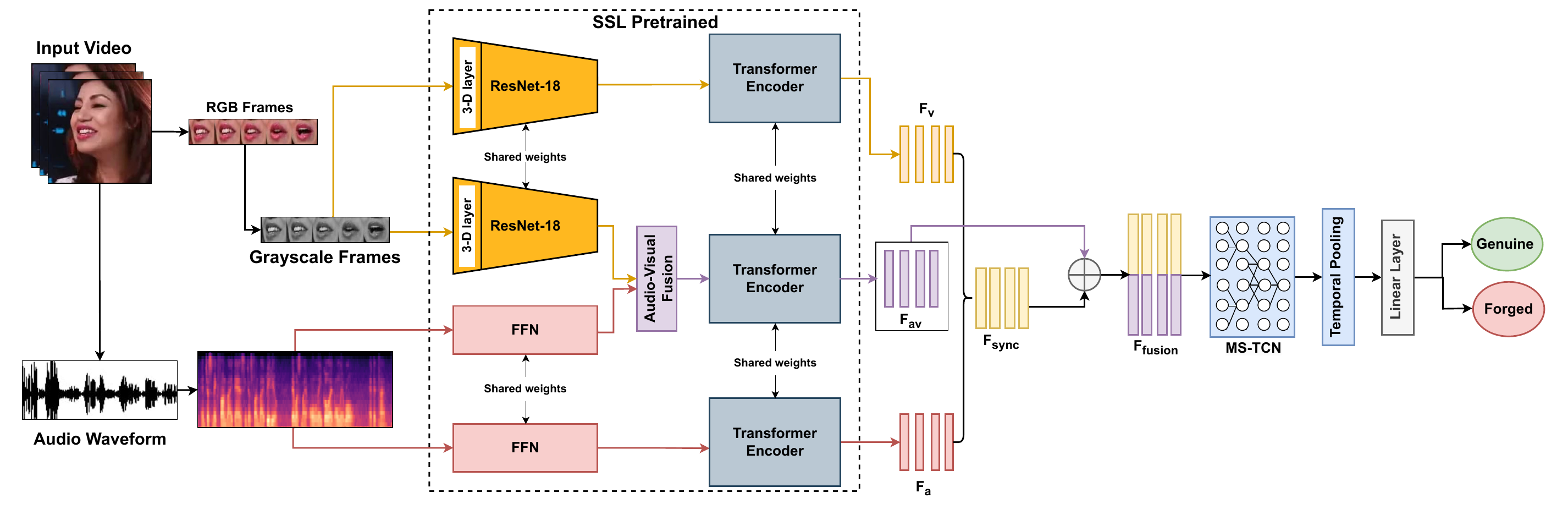}
  \vspace{-6mm}
  \caption{
  The proposed AV-Lip-Sync+ architecture for multimodal forgery detection. The lip image sequence is extracted from the input video, while the log filterbank energies are extracted from the audio track. The SSL pre-trained model consists of ResNet-18 for visual feature extraction, FFN for acoustic feature extraction, and a transformer encoder to extract spatiotemporal information from the visual and acoustic features. The extracted audio-visual features are further mapped through multi-scale temporal convolution network (MS-TCN), temporal pooling, and linear layer for classification.
  }
  \label{fig:proposed_model}
  \vspace{-4mm}
\end{figure*}


\section{Methodology}

As shown in Fig. \ref{fig:proposed_model}, our proposed AV-Lip-Sync+ model consist of three feature extractors, namely a lip image feature extractor, an acoustic feature extractor, and an audio-visual feature extractor. These feature extraction modules are followed by the Sync-Check Module, Feature Fusion Module, and Temporal Convolutional Network to capture the temporal correlation between visual and audio features. Finally, a temporal pooling layer and a linear layer are used for classification. Our feature extractors are based on the AV-HuBERT model~\cite{shi2021learning} pre-trained on the LRS3 dataset~\cite{afouras2018lrs3}. AV-HuBERT will be fine-tuned when training the detection model on the multimodal deepfake datasets.
\subsection{Audio-Visual Feature Extractor} As shown in Fig. \ref{fig:proposed_model}, the audio-visual feature extractor consists of a Resnet-18, a light-weight Feed Forward Network (FFN), and a transformer encoder. The 2-D Resnet-18 with front-end 3-D convolutional layers is used to extract lip-based visual features from each input lip image frame. The FFN is used to extract frame-level acoustic features from the input log filterbank energies of the audio waveform. These frame-level visual and acoustic features are concatenated along the feature dimension and fed to the transformer encoder, which generates a 768-D audio-visual embedding sequence $\vec F_{av}$ via
\begin{equation}\label{R3}
\vec F_{av} = {F_{\theta_e}(v, a)}, 
\end{equation}
where $v$ and $a$ represent frame-level visual features and acoustic features, respectively.


\subsection{Lip Image Feature Extractor}
In the lip image feature extractor, the output of Resnet-18 is fed to the transformer encoder, which generates the lip image embedding sequence $\vec F_{v}$ via
\begin{equation}\label{R1}
\vec F_{v} = {F_{\theta_e}(v, a_{dropout})},
\end{equation}
where $a_{dropout}$ indicates audio dropout, i.e., the audio information is not used.


\subsection{Acoustic Feature Extractor} 
In the acoustic feature extractor, the output of FFN is fed to the transformer encoder, which generates the acoustic feature embedding sequence $\vec F_{a}$ via
\begin{equation}\label{R2}
\vec F_{a} = {F_{\theta_e}(v_{dropout}, a)},
\end{equation}
where $v_{dropout}$ indicates video dropout, i.e., the visual information is not used.


\subsection{Sync-Check Module}
\label{sec:Sync-Check Module}
Because deepfake technology is not yet mature enough to generate synchronized and perfect audio-visual deepfakes, there is often a disharmony between the visual and audio modalities of deepfake videos, as shown in Fig. S1 in the Supplementary Material. This idea motivates us to exploit synchronization between lip movements and speech to detect forgeries in deepfake videos. Therefore, the input of the sync-check module includes the output representation ${\vec F_{v}}$ of the lip image feature extractor and the output representation ${\vec F_{a}}$ of the acoustic feature extractor. For each time frame $i$, we calculate the absolute difference between the corresponding lip embedding ${\vec F_{vi}}$ and audio embedding ${\vec F_{ai}}$ to capture the frame-level difference between the visual and acoustic modalities. Consequently, the output of the sync-check module is the sync-based feature vector sequence ${\vec F_{sync}}$ calculated as 
\begin{equation}\label{R4}
\vec F_{sync} = \{|{\vec F_{vi}} - {\vec F_{ai}}|\}_{i=1}^T,
\end{equation}
where $T$ is the number of frames in the input video.

\subsection{Feature Fusion Module} In addition to the sync-based feature vector sequence ${\vec F_{sync}}$, the robust audio-visual feature vector sequence $F_{av}$ obtained from the audio-visual feature extractor also captures the correlation between the two modalities. Therefore, we combine these two audio-visual representations for multimodal forgery detection. In the feature fusion module, $\vec F_{av}$ and $\vec F_{sync}$ are concatenated along the feature dimension to form a fusion representation sequence $\vec F_{fusion}$ as
\begin{equation}\label{fusion}
\vec F_{fusion} = {\vec F_{av} \oplus \vec F_{sync}},
\end{equation}
where $\oplus$ denotes the concatenation operation.

\subsection{Temporal Convolutional Network and Classifier} The temporal dynamics across the audio and visual frames contain important information about the video content. To capture inter-modal and intra-modal temporal correlations, we adopt the multi-scale temporal convolutional network (MS-TCN) in~\cite{martinez2020lipreading}. Temporal convolution takes a sequence of frame-level feature vectors and maps them into another sequence of the same dimension using one-dimensional temporal convolution. The temporal convolutional network acts as a sequence encoder, capturing short-term and long-term information by providing the network with visibility across multiple time scales. MS-TCN is followed by a temporal pooling layer and a linear layer for outputting the Real/Fake probability given $\vec F_{fusion}$ as
\begin{equation}
    \hat{y} = {F_{{\theta}_m} (\vec F_{fusion})},
 \end{equation}
where $\hat{y}$ represents the probability of the target class.


\subsection{Model Training}
The model is trained with the cross-entropy loss, defined as
\begin{equation}
    {L(y, \hat{y})=-\frac{1}{N} \sum_{i=1}^{N} [y_{i} \log \hat{y}_{i}+\left(1-y_{i}\right) \log \left(1-\hat{y}_{i}\right)]},
 \label{eq:ce_loss}
 \end{equation}
 where $N$ is the number of training samples, $y_{i}$ represents the ground truth label of the $i$-th sample (0 or 1), and $\hat{y}_{i}$ represents the class prediction probability of the $i$-th sample.
During training, the pre-trained front-end feature extractors and transformer encoder are fine-tuned, while MS-TCN and the linear classifier are trained from scratch.

\begin{table*}[htbp]
\caption{Statistics of Multimodal Fogery Datasets for Deepfake Detection.}
\vspace{-2mm}
    \begin{tabular}{c|c|c|c|c|c|c}
    \hline
         \textbf{Datasets} & \textbf{Real Videos} & \textbf{Fake Videos} & \textbf{Manipulation Methods} & \textbf{No of Subjects} & \textbf{Visual Manipulation} &  \textbf{Audio Manipulation}  \\ 
         \hline\hline
          \begin{tabular}{@{}c@{}}FakeAVCeleb \cite{khalid2021fakeavceleb} \end{tabular}   & {500} & {20000} & \begin{tabular}{@{}c@{}}Faceswap, Fsgan, \\ wav2lip, RTVC \end{tabular} &{500} & {Yes} & {Yes} \\ 
          \hline
          \begin{tabular}{@{}c@{}}DeepfakeTIMIT \cite{korshunov2018deepfakes} \end{tabular} & 320 & 320 & Faceswap & 32 & Yes & No \\ 
          \hline
    \end{tabular}
    \label{tab:Datasets}
    \vspace{-4mm}
\end{table*}

\subsection{AV-Lip-Sync+ with FE}
 As reported in~\cite{shahzad2022lip}, speech-lip synchronization based methods may not be good at detecting fake videos generated by some visual manipulation models such as Faceswap and FS-GAN. In these types of fake videos, the video contains only visual manipulation, while the audio is genuine. Furthermore, the manipulation does not necessarily occur in the lip region, but artifacts can be observed in other regions or throughout the face, including face boundaries. Deepfake detectors based on lip features may fail when the lip region is not manipulated, or when there is little manipulation in the mouth region. To address these issues, we use a face encoder to utilize the entire face features to enhance our proposed deepfake detector and make it more robust and generalizable to deep face manipulation techniques. To this end, we employ the pre-trained ViViT model~\cite{arnab2021vivit} as the face encoder to extract the spatiotemporal face features. Using tubelet embeddings and spatial and temporal transformers, the face encoder extracts inter- and intra-frame information from video content. The output of the face encoder is a single-vector representation, which is fed to a linear layer for classification. The face-based deepfake detection model is trained on multimodal deepfake dataset using cross-entropy loss. During training, the pre-trained face encoder is fine-tuned, while the linear classifier is trained from scratch.
 
 The model that combines AV-Lip-Sync+ and the face encoder is called AV-Lip-Sync+ with FE. The extracted single-vector representation of the face encoder and the representation obtained from the AV-Lip-Sync+ model are concatenated and fed to a two-layer linear classifier. During training, the pre-trained face encoder and AV-Lip-Sync+ are fixed, and only the classifier is trained using cross-entropy loss.

\section{Experiments}
We conducted experiments on two datasets: FakeAVCeleb~\cite{khalid2021fakeavceleb} and DeepfakeTIMIT~\cite{korshunov2018deepfakes}. Unlike other unimodal audio or video deepfake datasets, these two datasets contain both audio and visual modalities, and their fake samples contain audio and/or visual manipulations. Furthermore, the faces in the videos in both datasets are frontal, which makes them suitable for lip frame extraction as visual input to the proposed model.
The statistics of the two dataset are shown in Table \ref{tab:Datasets}.

\subsection{Datasets}
\subsubsection{FakeAVCeleb} The FakeAVCeleb dataset is an audio-visual dataset released in 2021 specifically designed for the deepfake detection task. It is based on a collection of 500 YouTube videos featuring 500 celebrities from diverse ethnic regions including South Asia, East Asia, Africa, Europe and America. Fake videos are generated from these 500 real videos using the Faceswap \cite{korshunova2017fast}, Fsgan \cite{nirkin2019fsgan}, wav2lip \cite{prajwal2020lip}, and real-time voice cloning (SV2TTS) \cite{jia2018transfer} manipulation methods and their combinations. Several examples are shown in Fig. \ref{fig_1}. In the case of Faceswap-wav2lip, the video is manipulated using both Faceswap and wav2lip manipulation methods. Similarly, in the case of Fsgan-wav2lip, the video is generated using a combination of Fsgan and wav2lip. The videos manipulated by wav2lip include two types, namely Fake-Video-Real-Audio (FVRA) and Fake-Video-Fake-Audio (FVFA). Wav2lip FVRA videos contain manipulated lips and real audio. In the case of wav2lip FVFA, in addition to lip manipulation, a real-time voice cloning method is also used to manipulate the audio. In total, the dataset contains 500 real videos and more than 20000 forged videos. 

Following \cite{shahzad2022lip, hashmi2022multimodal}, we used multiple test sets, namely Faceswap, Fsgan, RTVC, wav2lip, Faceswap-wav2lip, and Fsgan-wav2lip. Furthermore, two other major and diverse test sets are Test-set-1 and Test-set-2. In Test-set-1, the number of samples is the same for all manipulation methods, while in Test-set-2, the number of samples of RVFA (Real-Video-Fake-Audio), FVRA (Fake-Video-Real-Audio), and FVFA (Fake-Video-Fake-Audio) in the fake class is the same. The training-test split is based on the number of subjects in the dataset. The training set and test set contain real and fake videos corresponding to 430 subjects and 70 subjects respectively. Furthermore, all the test sets are balanced in terms of real and fake videos and contain 70 videos per class (real and fake). The training set contains only 430 real videos, which is significantly less than the number of fake videos. If the imbalance problem is not properly resolved, the experimental results will be biased. To eliminate the imbalance problem, we took real videos from the VoxCeleb1 dataset~\cite{Nagrani17} to make the training data of the real and fake classes more balanced and employed a more effective sampling method to rebalance class distributions during model training.



\subsubsection{DeepfakeTIMIT}
The DeepfakeTIMIT dataset contains 320 audio-visual human speech recordings from 32 subjects and is a subset of the VidTIMIT dataset \cite{sanderson2002vidtimit}. Each subject has 10 videos. 
For each video, the corresponding fake video is generated by the Faceswap manipulation method. The audio in both real and fake videos is always real. Since the dataset is small, we performed 5-fold cross-validation on it and evaluated the average performance.

\subsection{Preprocessing}
Our model mainly utilizes lip and audio features for multimodal forgery detection. For this purpose, the initial step is to extract the lip region from the frontal face using facial landmarks. We leveraged a pre-trained CNN-based face detector from the Dlib toolkit~\cite{king2009dlib}. The lip image sequence extracted from the input video is $96 \times 96$ RGB pixels. Before the lip image sequences are fed into the model, they are converted to grayscale. The input shape of the extracted lip features is $C \times F \times H \times W$, where $C$ represents the number of channels, $F$ denotes the number of frames, and $H$ and $W$ represent the height and width of each frame, respectively. In addition, for the audio modality, the waveform is extracted from the video and then converted to the log filterbank energies as the acoustic input of the model.
\par For facial feature extraction, among the different variants of ViViT in~\cite{arnab2021vivit}, we selected the best performing factorized encoder model as the face encoder. The visual encoder has two transformer blocks, namely a spatial transformer and a temporal transformer. The input to the model is short fixed-length video clips from the entire video. The number of frames in a clip is $16$, the input frame size is $224 \times 224$, the patch size is $16$, the number of input channels is $3$, and the embedding dimension is $768$. We used the tubelet embedding method. The ViViT model is pre-trained on the kinetics dataset~\cite{kay2017kinetics}. 
\begin{table}[t!]
\caption{Evaluation results of AV-Lip-Sync+ on the FakeAVCeleb dataset (in \%).}
\vspace{-2mm}
\centering
\scalebox{0.90}{
    \begin{tabular}{c|c|c|c|c|c}
    \hline
         \textbf{Test set} & \textbf{Class} & \textbf{Precision} & \textbf{Recall} & \textbf{F1-score} & \textbf{Accuracy} \\ 
         \hline\hline
          \multirow{2}{*}{Faceswap} & {Real} & 85.19 & 98.57 & 91.39 & \multirow{2}{*}{90.71} \\ 
          \cline{2-5} & {Fake} & 98.31 & 82.86 & 89.92 & \\
          \hline
          \multirow{2}{*}{Faceswap\_wav2lip} & {Real} & 100.0 & 98.57 & 99.28 & \multirow{2}{*}{99.29} \\ 
          \cline{2-5} & {Fake} & 98.59 & 100.0 & 99.29 &\\
          \hline
           \multirow{2}{*}{Fsgan} & {Real} & 86.25 & 99.57 & 92.00 & \multirow{2}{*}{91.43} \\ 
          \cline{2-5} & {Fake} & 98.33 & 84.29 & 90.77 &\\
          \hline
          \multirow{2}{*}{Fsgan\_wav2lip} & {Real} & 100.0 & 98.57 & 99.28 & \multirow{2}{*}{99.29} \\
          \cline{2-5} & {Fake} & 98.59 & 100.0 & 99.29 & \\
          \hline
          \multirow{2}{*}{RTVC} & {Real} & 95.83 & 98.57 & 97.18 & \multirow{2}{*}{97.14}  \\ 
          \cline{2-5} & {Fake} & 98.53 & 95.71 & 97.10 &\\
          \hline
          \multirow{2}{*}{Wav2lip} & {Real} &  100.0 & 98.57 & 99.28 & \multirow{2}{*}{99.29} \\ 
          \cline{2-5} & {Fake} & 98.59 & 100.0 & 99.29 &\\
          \hline
          \multirow{2}{*}{Test-set-1} & {Real} & 93.24 & 98.57 & 95.83 & \multirow{2}{*}{95.71} \\ 
          \cline{2-5} & {Fake} & 98.48 & 92.86 & 95.59 &\\
          \hline
          \multirow{2}{*}{Test-set-2} & 
          {Real} & 98.57 & 98.57 & 98.57 & \multirow{2}{*}{98.57} \\ 
          \cline{2-5} &{Fake} & 98.57 & 98.57 & 98.57 & \\
          \hline
    \end{tabular}}
    \label{tab:AV_HuBERT_on_all_test_sets}
    \vspace{-4mm}
\end{table}

\subsection{Model Configuration and Training}
Our proposed AV-Lip-Sync+ model contains 132.85M trainable parameters. When incorporating the Face Encoder (FE) module, the parameters increase to 249.37M due to the additional fusion mechanism. To evaluate the inference time, we conducted experiments on an NVIDIA GeForce RTX 2080 Ti GPU. To ensure reliable measurements, the model was executed in evaluation mode, and multiple warm-up iterations were performed to mitigate cold-start overhead. The average inference time per video sample is 0.07 seconds for AV-Lip-Sync+ and 0.14 seconds for AV-Lip-Sync+ with FE. These measurements only consider the forward pass of the model and do not include the time for pre- or post-processing steps.

Our model was trained by the Adam optimizer with a learning rate of 0.00001 and an early stopping strategy for $30$ epochs. We added 3570 real videos from the VoxCeleb1 dataset~\cite{Nagrani17} to the real class and employed a more effective sampling method called Imbalanced Dataset Sampler. This method rebalances class distributions during model training by ensuring that each batch contains a balanced number of samples from both classes (real and fake) and automatically estimates the sampling weights. Through this strategy, we not only address the challenge of dataset imbalance, but also overcome the overfitting and information loss issues associated with traditional over-sampling and under-sampling. 

The evaluation metrics used include precision, recall, F1-score, accuracy (see Section S2 in the Supplementary Material for details), the receiver operating characteristics (ROC) curve, and the area under the curve (AUC). We report video-level performance rather than frame-level performance.

\begin{figure}[]
\vspace{-1mm}
  \includegraphics[width=0.45\textwidth]{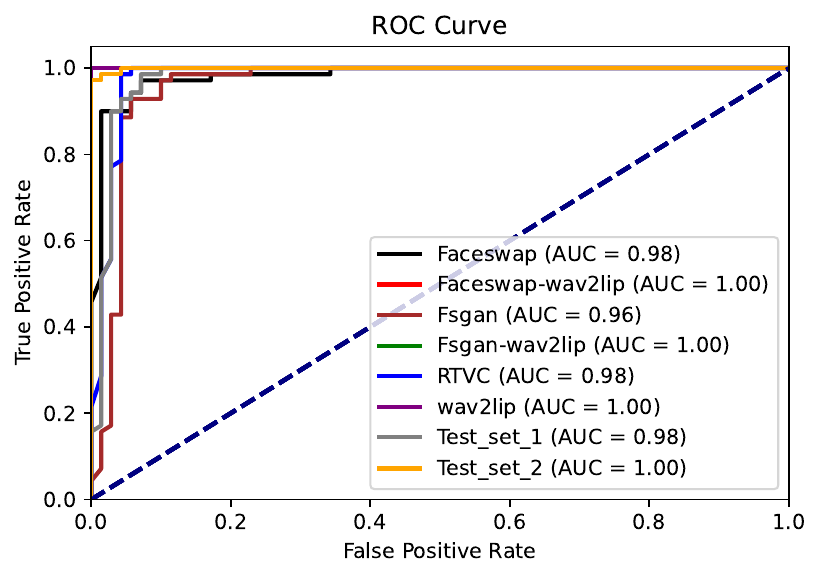}
  \vspace{-3mm}
  \caption{ROC curves and AUC scores of the proposed AV-Lip-Sync+ method on various test sets of the FakeAVCeleb dataset.}
  \label{fig:auc_roc}
  \vspace{-4mm}
\end{figure}


\subsection{Results}
\subsubsection{Evaluation of AV-Lip-Sync+ and AV-Lip-Sync+ with FE} Table \ref{tab:AV_HuBERT_on_all_test_sets} shows the performance of the proposed AV-Lip-Sync+ model evaluated on the FakeAVCeleb dataset. The manipulation methods of Faceswap, Faceswap-wav2lip, Fsgan, Fsgan-wav2lip, RTVC, and wav2lip are all seen during the model training process. It is obvious that AV-Lip-Sync+ performed well on almost all test sets except Faceswap and Fsgan. The main reason may be that the AV-HuBERT feature extractor only extracts visual information from lip images and ignores information outside the lip region. 
Fig. 3 shows the ROC curves and AUC scores for all test sets. The AUC scores under different manipulation conditions are all above 0.96.

Facial features are crucial for detecting deepfake videos, as local or entire face regions may contain artifacts caused by video manipulation. To address the limitations of AV-Lip-Sync+ on Faceswap and Fsgan test samples, we used a ViViT-based face encoder to inject face embeddings into an ensemble model (called Lip-Sync+ with FE). As shown in Table \ref{tab:AVH_TCN_Fusion_with_other_models}, the accuracy of Faceswap and Fsgan test sets increased from 90.71\% and 91.43\% to 97.14\% and 96.43\%, respectively. The results show that providing supplementary information to the proposed model can improve the detection of videos with tampered faces through Faceswap and Fsgan techniques.

\begin{table*}[t]
\centering
\caption{Accuracy of AV-Lip-Sync+ and AV-Lip-Sync+ with FE on the FakeAVCeleb dataset (in \%).}
\vspace{-2mm}    
    \scalebox{1.0}{ 
    \begin{tabular}{c|c|c|c|c|c|c|c|c}
    \hline
         \textbf{Model} & \textbf{Faceswap} & \textbf{Faceswap\_wav2lip} & \textbf{Fsgan} & \textbf{Fsgan\_wav2lip} & \textbf{RTVC} & \textbf{Wav2lip} &  \textbf{Test-set-1} & \textbf{Test-set-2} \\
         \hline\hline
          \begin{tabular}{@{}c@{}}AV-Lip-Sync+ \end{tabular}   &  90.71 & \textbf{99.29} & {91.43} & \textbf{99.29} & \textbf{97.14} & \textbf{99.29} & 95.71 & 98.57 \\
          \hline
          \begin{tabular}{@{}c@{}}AV-Lip-Sync+ with FE \end{tabular}   &  \textbf{97.86} & \textbf{99.29} & \textbf{96.43} & \textbf{99.29} & 96.43 & \textbf{99.29} & \textbf{97.86} & \textbf{99.29} \\
          \hline
    \end{tabular}
    }
    \label{tab:AVH_TCN_Fusion_with_other_models}
    \vspace{-4mm}
\end{table*}

\begin{table*}[t]
\centering
\caption{Evaluation results of different models on the FakeAVCeleb dataset (in \%).}
\vspace{-2mm}    
    \scalebox{0.90}{ 
    \begin{tabular}{c|c|c|c|c|c|c|c}
    \hline
         \textbf{Type} & \textbf{Model} &  \textbf{Modality} & \textbf{Class} & \textbf{Precision} & \textbf{Recall} & \textbf{F1-score} & \textbf{Accuracy}\\
         \hline\hline
          \multirow{2}{*}{Unimodal\cite{khalid2021evaluation}} & \multirow{2}{*}{VGG16} & \multirow{2}{*}{V} & {Real} & 69.35 & 89.66 & 78.21 &  \multirow{2}{*}{81.03} \\
          \cline{4-7} &&& {Fake} & 87.24 & 77.50 & 82.08 & \\
          \hline
          \multirow{2}{*}{Unimodal \cite{khalid2021evaluation}} & \multirow{2}{*}{Xception} & \multirow{2}{*}{A} & {Real} & 87.50 & 60.87 & 71.79 & \multirow{2}{*}{76.26} \\
          \cline{4-7} &&& {Fake} & 70.33 & 91.43 & 79.50 & \\
          \hline
          \multirow{2}{*}{Unimodal \cite{haliassos2021lips}} & \multirow{2}{*}{LipForensics} & \multirow{2}{*}{V} & {Real} & 70.00 & 91.00 & 80.00 & \multirow{2}{*}{76.00} \\
          \cline{4-7} &&& {Fake} & 88.00 & 61.00 & 72.00 & \\
          \hline
           \multirow{2}{*}{Ensemble (Soft-Voting)\cite{khalid2021evaluation}} & \multirow{2}{*}{VGG16} & \multirow{2}{*}{AV} & {Real} & 69.35 & 89.66 & 78.21 & \multirow{2}{*}{78.04} \\
          \cline{4-7} &&& {Fake} & 89.48 & 68.94 & 77.88 & \\
          \hline
           \multirow{2}{*}{Ensemble (Hard-Voting)\cite{khalid2021evaluation}} & \multirow{2}{*}{VGG16} & \multirow{2}{*}{AV} & {Real} & 69.35 & 89.66 & 78.21 & \multirow{2}{*}{78.04} \\
          \cline{4-7} &&& {Fake} & 89.48 & 68.94 & 77.88 & \\
          \hline
          \multirow{2}{*}{Multimodal-1\cite{khalid2021evaluation}} & \multirow{2}{*}{Multimodal-1} & \multirow{2}{*}{AV} & {Real} & 00.00 & 00.00 & 00.00& \multirow{2}{*}{50.00} \\
          \cline{4-7} &&& {Fake} & 49.60 & 100.0 & 66.30 & \\ 
          \hline
          \multirow{2}{*}{Multimodal-2\cite{khalid2021evaluation}} & \multirow{2}{*}{Multimodal-2} & \multirow{2}{*}{AV} & {Real} & 71.00 & 58.70 & 64.30 & \multirow{2}{*}{67.40} \\
          \cline{4-7} &&& {Fake} & 64.80 & 76.00 & 70.00 & \\
          \hline
          \multirow{2}{*}{Multimodal-3\cite{khalid2021evaluation}} & \multirow{2}{*}{CDCN} & \multirow{2}{*}{AV} & {Real} & 50.00 & 06.80 & 12.00 & \multirow{2}{*}{51.50} \\
          \cline{4-7} &&& {Fake} & 50.00 & 94.00 & 65.10 & \\
          \hline
          \multirow{2}{*}{Multimodal-4\cite{chugh2020not}} & \multirow{2}{*}{MDS} & \multirow{2}{*}{AV} & {Real} & 62.16 & 98.57 & 76.24 & \multirow{2}{*}{69.29} \\
          \cline{4-7} &&& {Fake} & 96.55 &40.00 & 56.57 & \\
          \hline
          \multirow{2}{*}{{Multimodal-Ensemble\cite{hashmi2022multimodal}}} & \multirow{2}{*}{{Ensemble model}} & \multirow{2}{*}{{AV}} & {Real} &  {83.13} &  {98.57} &  {90.20} & \multirow{2}{*}{ {89.29}} \\
          \cline{4-7} &&& { {Fake}} &  {98.25} &  {80.00} &  {88.19} & \\
          \hline
          \multirow{2}{*}{{Multimodal\cite{raza2023multimodaltrace}}} & \multirow{2}{*}{{Multimodaltrace}} & \multirow{2}{*}{{AV}} & {Real} &  {-} &  {-} &  {-} & \multirow{2}{*}{ {92.90}} \\
          \cline{4-7} &&& {Fake} &  {-} &  {-} &  {-} & \\
          \hline
          \multirow{2}{*}{Multimodal-Ensemble\cite{ilyas2023avfakenet}} & \multirow{2}{*}{AVFakeNet} & \multirow{2}{*}{AV} & {Real} &  {-} & {-} & {-} & \multirow{2}{*}{93.41} \\
          \cline{4-7} &&& {Fake} &  {-} &  {-} & {-} & \\
          \hline
          \multirow{2}{*}{ Fusion\cite{yang2023avoid}} & \multirow{2}{*}{AVoiD-DF} & \multirow{2}{*}{AV} & {Real} & {-} & {-} & {-} & \multirow{2}{*}{83.70} \\
          \cline{4-7} &&& {Fake} &  {-} &  {-} & {-} & \\
           \hline
          \multirow{2}{*}{ Fusion\cite{katamneni2023mis}} & \multirow{2}{*}{MIS-AVioDD } & \multirow{2}{*}{AV} & {Real} & {-} & {-} & {-} & \multirow{2}{*}{96.20} \\
          \cline{4-7} &&& {Fake} &  {-} &  {-} & {-} & \\
          \hline
          
          \multirow{2}{*}{{Multimodal\cite{shahzad2022lip}}} & \multirow{2}{*}{{AV-Lip-Sync}} & \multirow{2}{*}{{AV}} & {Real} &  {91.78} &  {95.71} &  {93.71} & \multirow{2}{*}{ {93.57}} \\
          \cline{4-7} &&& { {Fake}} &  {95.52} &  {91.43} &  {93.43} & \\
          \hline

           \multirow{2}{*}{{Identity Aware\cite{cozzolino2023audio}}} & \multirow{2}{*}{{POI-Forensics}} &  \multirow{2}{*}{{AV}} &  {Real} &  {-} &   {-} &   {-} &  \multirow{2}{*}{{85.50}} \\ \cline{4-7}  &&& {Fake} &   {-} &  {-} & {-} & \\
            \hline
            
            \multirow{2}{*}{{Contrastive Learning\cite{liu2023mcl}}} & \multirow{2}{*}{{MCL}} &  \multirow{2}{*}{{AV}} &  {Real} &  {-} &   {-} &   {-} &  \multirow{2}{*}{{89.25}} \\ \cline{4-7}  &&& {Fake} &   {-} &  {-} & {-} & \\
            \hline
            
            \multirow{2}{*}{{Multimodal\cite{yu2023pvass}}} & \multirow{2}{*}{{PVASS-MDD}} &  \multirow{2}{*}{{AV}} &  {Real} &  {-} &   {-} &   {-} &  \multirow{2}{*}{{95.70}} \\ \cline{4-7}  &&& {Fake} &   {-} &  {-} & {-} & \\
            \hline
            
            \multirow{2}{*}{{Multimodal\cite{feng2023self}}} & \multirow{2}{*}{{A-V Anomaly}} &  \multirow{2}{*}{{AV}} &  {Real} &  {-} &   {-} &   {-} &  \multirow{2}{*}{{92.71}} \\ \cline{4-7}  &&& {Fake} &   {-} &  {-} & {-} & \\
            \hline
            
            \multirow{2}{*}{{Multimodal\cite{tian2023unsupervised}}} & \multirow{2}{*}{{Intra-Cross-modal}} &  \multirow{2}{*}{{AV}} &  {Real} &  {-} &   {-} &   {-} &  \multirow{2}{*}{{94.59}} \\ \cline{4-7}  &&& {Fake} &   {-} &  {-} & {-} & \\
            \hline
            
            \multirow{2}{*}{{Fusion\cite{li2024zero}}} & \multirow{2}{*}{{Zero-Shot}} &  \multirow{2}{*}{{AV}} &  {Real} &  {-} &   {-} &   {-} &  \multirow{2}{*}{{91.94}} \\ \cline{4-7}  &&& {Fake} &   {-} &  {-} & {-} & \\
            \hline
            
            \multirow{2}{*}{{Multimodal\cite{katamneni2024contextual}}} & \multirow{2}{*}{{MMMS-BA}} &  \multirow{2}{*}{{AV}} &  {Real} &  {-} &   {-} &   {-} &  \multirow{2}{*}{{97.90}} \\ \cline{4-7}  &&& {Fake} &   {-} &  {-} & {-} & \\
            \hline
            
            \multirow{2}{*}{{Multimodal\cite{nie2024frade}}} & \multirow{2}{*}{{FRADE}} &  \multirow{2}{*}{{AV}} &  {Real} &  {-} &   {-} &   {-} &  \multirow{2}{*}{{98.60}} \\ \cline{4-7}  &&& {Fake} &   {-} &  {-} & {-} & \\
            \hline
            
            \multirow{2}{*}{{Multimodal\cite{oorloff2024avff}}} & \multirow{2}{*}{{AVFF}} &  \multirow{2}{*}{{AV}} &  {Real} &  {-} &   {-} &   {-} &  \multirow{2}{*}{{98.60}} \\ \cline{4-7}  &&& {Fake} &   {-} &  {-} & {-} & \\
            \hline

            \multirow{2}{*}{{Multimodal-Ensemble\cite{hashmi2023avtenet}}} & \multirow{2}{*}{{AVTENet}} & \multirow{2}{*}{{AV}} & {Real} &  {100.0} &  {97.14} &  {98.55} & \multirow{2}{*}{ {98.57}} \\
              \cline{4-7} &&& { {Fake}} &  {97.22} &  {100.0} &  {98.59} & \\
              \hline
               
          \multirow{2}{*}{\textbf{Multimodal (ours)}} & \multirow{2}{*}{\textbf{ AV-Lip-Sync+}} & \multirow{2}{*}{\textbf{AV}} & \textbf{Real} & \textbf {98.57} & \textbf {98.57} & \textbf {98.57} & \multirow{2}{*}{\textbf {98.57}} \\
          \cline{4-7} &&& {\textbf {Fake}} & \textbf {98.57} & \textbf {98.57} & \textbf {98.57} & \\
          \hline
          \multirow{2}{*}{\textbf{Multimodal (ours)}} & \multirow{2}{*}{\textbf{AV-Lip-Sync+ with FE}} & \multirow{2}{*}{\textbf{AV}} & \textbf{Real} & \textbf {100.0} & \textbf {98.57} & \textbf {99.28} & \multirow{2}{*}{\textbf {99.29}} \\
          \cline{4-7} &&& {\textbf {Fake}} & \textbf {98.59} & \textbf {100.0} & \textbf {99.29} & \\
          \hline
    \end{tabular}
    }
    \label{tab:Final}
    \vspace{-4mm}
\end{table*}

\subsubsection{Comparison of different models} In this experiment, we compared our models with various existing unimodal, multimodal, fusion, and ensemble deepfake detection models on Test-set-2 of the FakeAVCeleb dataset. The results are shown in Table \ref{tab:Final}. Several unimodal, multimodal and ensemble models have been evaluated in~\cite{khalid2021evaluation}, but the performance of most models is unsatisfactory.Unimodal video-only models VGG16 and LipForensics~\cite{haliassos2021lips} rely solely on visual features for deepfake detection. VGG16 processes video frames, while LipForensics focuses on lip sequences. Similarly, the unimodal audio-only model Xception is trained using MFCC features extracted from the audio modality. However, these unimodal models are doomed to fail in situations where the focusing modality is genuine but the other modality is manipulated.

The ensemble and multimodal models evaluated in~\cite{khalid2021evaluation}, which aggregate the predictions of unimodal classifiers through simple voting without exploiting inter-modal relationships, or were not originally designed for forgery detection (e.g., Multimodal-1, Multimodal-2, and CDCN), did not lead to performance improvements compared to unimodal deepfake detectors. MDS~\cite{chugh2020not}, a modality dissonance score-based deepfake detector, also only achieved 69\% accuracy.

Later, multimodal ensemble models~\cite{hashmi2022multimodal, raza2023multimodaltrace, ilyas2023avfakenet} outperformed unimodal and early multimodal or ensemble models by combining more effective sub-modules. 
The identity-based POI-Forensic model~\cite{cozzolino2023audio} achieves an accuracy of 85.50\%, which is constrained by the identities used during training, limiting its applicability in real-world scenarios, such as detecting deepfakes in user-generated content on social media. The Multimodal Contrastive Learning (MCL) method~\cite{liu2023mcl} utilizes contrastive learning to bridge the cross-modal gap and achieves an accuracy of 89.25\%. PVASS-MDD~\cite{yu2023pvass} employs a two-stage framework: a self-supervised module aligns visual-audio features by predicting audio from visual cues, and a detection stage leverages this alignment to enhance detection of audio-visual inconsistencies in deepfake videos. A-V Anomaly method~\cite{feng2023self} trains an autoregressive model on real unlabeled videos, detecting manipulated videos with low generation probability, achieving an accuracy of 92.71\%. However, it is less effective against manipulations that preserve synchronization, such as those altering a person's appearance without altering mouth motion. An unsupervised method~\cite{tian2023unsupervised} detects deepfakes by spotting intra- and cross-modal inconsistencies, achieving 94.59\% accuracy. 
A zero-shot method~\cite{li2024zero} detects deepfakes by comparing  automatic speech recognition (ASR) and visual speech recognition (VSR) outputs via edit distance, achieving 91.94\% accuracy. MMMS-BA framework~\cite{katamneni2024contextual} applies attention over audio-visual sequences to address the modality gap and improve deepfake detection and localization, reaching 97.90\% accuracy.
The AV-Lip-Sync model~\cite{shahzad2022lip} uses speech-lip synchronization features based on the difference between the extracted lip sequence and the synthetic lip sequence and achieves an accuracy of 93.57\%. It requires the wav2lip~\cite{hegde2021visual} module to generate lip sequences from audio, which increases model complexity, training, and inference time.
Based on these previous studies, the current state-of-the-art models such as AVTENet~\cite{hashmi2023avtenet}, RADE~\cite{nie2024frade}, and AVFF~\cite{oorloff2024avff} have shown excellent performance with detection accuracies higher than 98\%. These models are based on advanced model architectures, more effective training strategies and cross-modal modeling, or large-scale pre-trained models.

Our proposed AV-Lip-Sync+ model eliminates the need for a wav2lip generator while achieving significant performance improvements compared to the AV-Lip-Sync model. AV-Lip-Sync+ and AV-Lip-Sync+ with FE achieve 98.57\% and 99.29\% accuracy, respectively. On Test-set-2 of the FakeAVCeleb dataset, AV-Lip-Sync+ achieves state-of-the-art performance, and AV-Lip-Sync+ with FE sets a new state-of-the-art benchmark. These results highlight the effectiveness of leveraging SSL features to model speech-lip synchronization, audio-visual correlations, and spatiotemporal facial artifacts for deepfake video detection.

Ablation studies as well as the discriminant analysis of features and experiments on robustness to partial occlusion and generalization across datasets are detailed in Sections S3, S4, S5, and S6 in the Supplementary Material, respectively.

\begin{table}[h!]
\caption{Evaluation results of AV-Lip-Sync+ on the DeepfakeTIMIT dataset (in \%).}
\vspace{-2mm}
\centering
\scalebox{0.90}{
    \begin{tabular}{c|c|c|c|c}
    \hline
     \textbf{Type} & \textbf{Model} &  \textbf{Modality} & \textbf{Quality} & \textbf{AUC}\\
     \hline\hline
      \multirow{2}{*}{Unimodal \cite{nguyen2019capsule}} & 
      \multirow{2}{*}{Capsule} & \multirow{2}{*}{V} & {LQ}  &78.40 \\
      \cline{4-5} &&& {HQ} & 74.40 \\
      \hline
      \multirow{2}{*}{Unimodal \cite{nguyen2019multi}} & 
      \multirow{2}{*}{Multi-task} & \multirow{2}{*}{V} & {LQ}  & 62.20 \\
      \cline{4-5} &&& {HQ} & 55.30 \\
      \hline
      \multirow{2}{*}{Unimodal \cite{yang2019exposing}} & 
      \multirow{2}{*}{HeadPose} & \multirow{2}{*}{V} & {LQ}  &55.10 \\
      \cline{4-5} &&& {HQ} & 53.20 \\
      \hline
      \multirow{2}{*}{Unimodal \cite{han2017two}} & 
      \multirow{2}{*}{Two-stream} & \multirow{2}{*}{V} & {LQ}  & 83.50 \\
      \cline{4-5} &&& {HQ} & 73.50 \\
      \hline
      \multirow{2}{*}{Unimodal \cite{matern2019exploiting}} & 
      \multirow{2}{*}{VA-MLP} & \multirow{2}{*}{V} & {LQ}  & 61.40 \\
      \cline{4-5} &&& {HQ} & 62.10 \\
      \hline
      \multirow{2}{*}{Unimodal  \cite{matern2019exploiting}} & 
      \multirow{2}{*}{VA-LogReg} & \multirow{2}{*}{V} & {LQ}  & 77.00 \\
      \cline{4-5} &&& {HQ} & 77.30 \\
      \hline
      \multirow{2}{*}{Unimodal \cite{afchar2018mesonet}} & 
      \multirow{2}{*}{Meso-4} & \multirow{2}{*}{V} & {LQ}  & 87.80 \\
      \cline{4-5} &&& {HQ} & 68.40 \\
      \hline
      \multirow{2}{*}{Unimodal \cite{rossler2019faceforensics++}} & 
      \multirow{2}{*}{Xception-raw} & \multirow{2}{*}{V} & {LQ}  & 56.70 \\
      \cline{4-5} &&& {HQ} & 54.00 \\
      \hline
      \multirow{2}{*}{Unimodal \cite{rossler2019faceforensics++}} & 
      \multirow{2}{*}{Xception-c40} & \multirow{2}{*}{V} & {LQ}  & 75.80 \\
      \cline{4-5} &&& {HQ} & 70.50 \\
      \hline
      \multirow{2}{*}{Unimodal \cite{rossler2019faceforensics++}} & 
      \multirow{2}{*}{Xception-c23} & \multirow{2}{*}{V} & {LQ}  & 95.90 \\
      \cline{4-5} &&& {HQ} & 94.40 \\
      \hline
      \multirow{2}{*}{Unimodal \cite{li2018exposing}} & 
      \multirow{2}{*}{FWA} & \multirow{2}{*}{V} & {LQ}  & 99.90 \\
      \cline{4-5} &&& {HQ} & 93.20 \\
      \hline
      \multirow{2}{*}{Unimodal \cite{li2018exposing}} & 
      \multirow{2}{*}{DSP-FWA} & \multirow{2}{*}{V} & {LQ}  & 99.90 \\
      \cline{4-5} &&& {HQ} & 99.70 \\
      \hline
       \multirow{2}{*}{Multimodal \cite{mittal2020emotions}} & 
      \multirow{2}{*}{Emotions don't lie} & \multirow{2}{*}{AV} & {LQ}  & 96.30 \\
      \cline{4-5} &&& {HQ} & 94.90 \\
      \hline
      \multirow{2}{*}{Multimodal \cite{chugh2020not}} & 
      \multirow{2}{*}{\begin{tabular}{@{}c@{}}MDS\end{tabular}} & \multirow{2}{*}{AV} & {LQ}  & 97.92 \\
      \cline{4-5} &&& {HQ} & 96.87 \\
      \hline
      \multirow{2}{*}{Multimodal \cite{shahzad2022lip}} & 
      \multirow{2}{*}{AV-Lip-Sync} & \multirow{2}{*}{AV} & {LQ}  & 97.90 \\
      \cline{4-5} &&& {HQ} & 96.80 \\
      \hline

    \multirow{2}{*}{Multimodal \cite{cozzolino2023audio}} & 
      \multirow{2}{*}{POI-Forensics} & \multirow{2}{*}{AV} & LQ  & 98.20 \\
      \cline{4-5} &&& HQ & 99.20 \\
      \hline

      \multirow{2}{*}{Multimodal \cite{liu2023mcl}} & 
      \multirow{2}{*}{MCL} & \multirow{2}{*}{AV} & LQ & 97.20 \\
      \cline{4-5} &&& HQ & 99.09 \\
      \hline

       \multirow{2}{*}{Multimodal (ours)} & 
      \multirow{2}{*}{AV-Lip-Sync+} & \multirow{2}{*}{AV} & {LQ}  & 95.80 \\
      \cline{4-5} &&& {HQ} & 98.80 \\
      \hline
      \multirow{2}{*} {\textbf{Multimodal (ours)}} & 
      \multirow{2}{*} {\textbf{\begin{tabular}{@{}c@{}}AV-Lip-Sync+ with FE\end{tabular}}} & \multirow{2}{*}{\textbf{AV}} & \textbf{LQ}  & \textbf{99.96} \\
      \cline{4-5} &&& \textbf{HQ} & \textbf{99.98} \\
      \hline
      
    \end{tabular}}
    \label{tab:DFTIMIT}
    \vspace{-4mm}
\end{table}

\subsubsection{Evaluation on the DeepfakeTIMIT dataset} The DeepfakeTIMIT dataset is primarily used for training and evaluating visual deepfake detectors, as it only contains visual manipulation.
It comes in two versions, Low-Quality (LQ) and High-Quality (HQ). For a fair comparison, we compared the proposed model with unimodal visual and multimodal audio-visual deepfake detectors. Additionally, we performed 5-fold cross-validation and reported the average AUC score. As can been seen from Table \ref{tab:DFTIMIT}, among unimodal visual detectors, FWA and DSP-FWA~\cite{li2018exposing} achieved the best AUC of 99.90 under the LQ condition, and DSP-FWA achieved the best AUC of 99.70 under the HQ condition. Although the five existing audio-visual detectors (Emotions Don't lie~\cite{mittal2020emotions}, MDS~\cite{chugh2020not}, AV-Lip-Sync~\cite{shahzad2022lip}, POI-Forensics~\cite{cozzolino2023audio}, and MCL~\cite{liu2023mcl}) are overall better than most unimodal visual detectors, their performance is worse than that of the best performing unimodal visual detector DSP-FWA. However, our multimodal AV-Lip-Sync+ with FE outperformed all models compared in the table. The AUC score is 99.96 for LQ and 99.98 for HQ. By appropriately integrating pre-trained AV-HuBERT and ViViT models for audio-visual feature extraction, our proposed model achieves state-of-the-art results on the DeepfakeTIMIT dataset.

\vspace{-1mm}
\section{Conclusions}
In this study, we have used AV-HuBERT and ViViT for the downstream task of audio-visual video forgery detection. We exploit the inconsistency between visual and audio modalities using the powerful audio-visual representation provided by AV-HuBERT, which is pre-trained using self-supervised learning. Since AV-HuBERT only extracts visual features from the lip region, which may not be sufficient to detect artifacts outside the lip region, we also adopt another transformer-based model ViViT to exploit facial features. Overall, our model jointly exploits SSL audio/visual/audio-visual representations, synchronization features, temporal correlation between lip image frames and audio, and spatiotemporal facial features to detect deepfakes. Experimental results on the FakeAVCeleb and DeepfakeTIMIT datasets show that our model outperforms all existing models and achieves new state-of-the-art performance on both datasets. In future work, we will aim to further improve the generalizability of deepfake detection techniques in multimodal settings. 

\section*{Acknowledgment}

This work was supported in part by the National Science and Technology Council, Taiwan, under Grants NSTC 111-2221-E-001-002, NSTC 113-2221-E-004-001-MY3, and NSTC 113-2221-E-004-006-MY.



\vspace{-2mm}
\bibliographystyle{elsarticle-num}
\bibliography{references}

\end{document}